\documentclass{article}
\usepackage{ijcai24}

\usepackage{times}
\usepackage{soul}
\usepackage{url}
\usepackage[hidelinks]{hyperref}
\usepackage[utf8]{inputenc}
\usepackage[small]{caption}
\usepackage{graphicx}
\usepackage{amsmath}
\usepackage{amsthm}
\usepackage{booktabs}
\usepackage{algorithm}
\usepackage{algorithmic}
\usepackage[switch]{lineno}

\usepackage{multirow,booktabs}
\usepackage{amsmath}

\usepackage[table]{xcolor}

\newcommand{\firstc}{{\cellcolor[RGB]{ 224, 174, 208 }}}
\newcommand{\secondc}{{\cellcolor[RGB]{212, 226, 212 }}}
\newcommand{\thirdc}{{\cellcolor[RGB]{   255, 229, 229 }}}
\newcommand{\diagonal}{{\cellcolor[RGB]{  217,217,217 }}}

\title{DesignProbe: A Graphic Design Benchmark \\for Multimodal Large Language Models}

\author{
Jieru Lin$^1$\thanks{Work was done during the first author's internship at Microsoft.}
\and
Danqing Huang$^2$\and
Tiejun Zhao$^1$\and
Dechen Zhan$^1$\And
Chin-Yew Lin$^2$\\
\affiliations
$^1$Harbin Institute of Technology, Harbin, China\\
$^2$Microsoft\\
\emails
hitjierulin@gmail.com,
\{dahua, cyl\}@microsoft.com,
\{tjzhao, dechen\}@hit.edu.cn
}

\begin{document}

\maketitle

\begin{abstract}
    A well-executed graphic design typically achieves harmony in two levels, from the fine-grained design elements (color, font and layout) to the overall design. This complexity makes the comprehension of graphic design challenging, for it needs the capability to both recognize the design elements and understand the design.
    With the rapid development of Multimodal Large Language Models (MLLMs), we establish the \textit{DesignProbe}, a benchmark to investigate the capability of MLLMs in design. Our benchmark includes eight tasks in total, across both the fine-grained element level and the overall design level. At design element level, we consider both the attribute recognition and semantic understanding tasks. At overall design level, we include style and metaphor. 9 MLLMs are tested and we apply GPT-4 as evaluator. 
    Besides, further experiments indicates that refining prompts can enhance the performance of MLLMs. We first rewrite the prompts by different LLMs and found increased performances appear in those who self-refined by their own LLMs. We then add extra task knowledge in two different ways (text descriptions and image examples), finding that adding images boost much more performance over texts.
\end{abstract}
\begin{figure}[htb]
\centering
\includegraphics[scale=0.67]{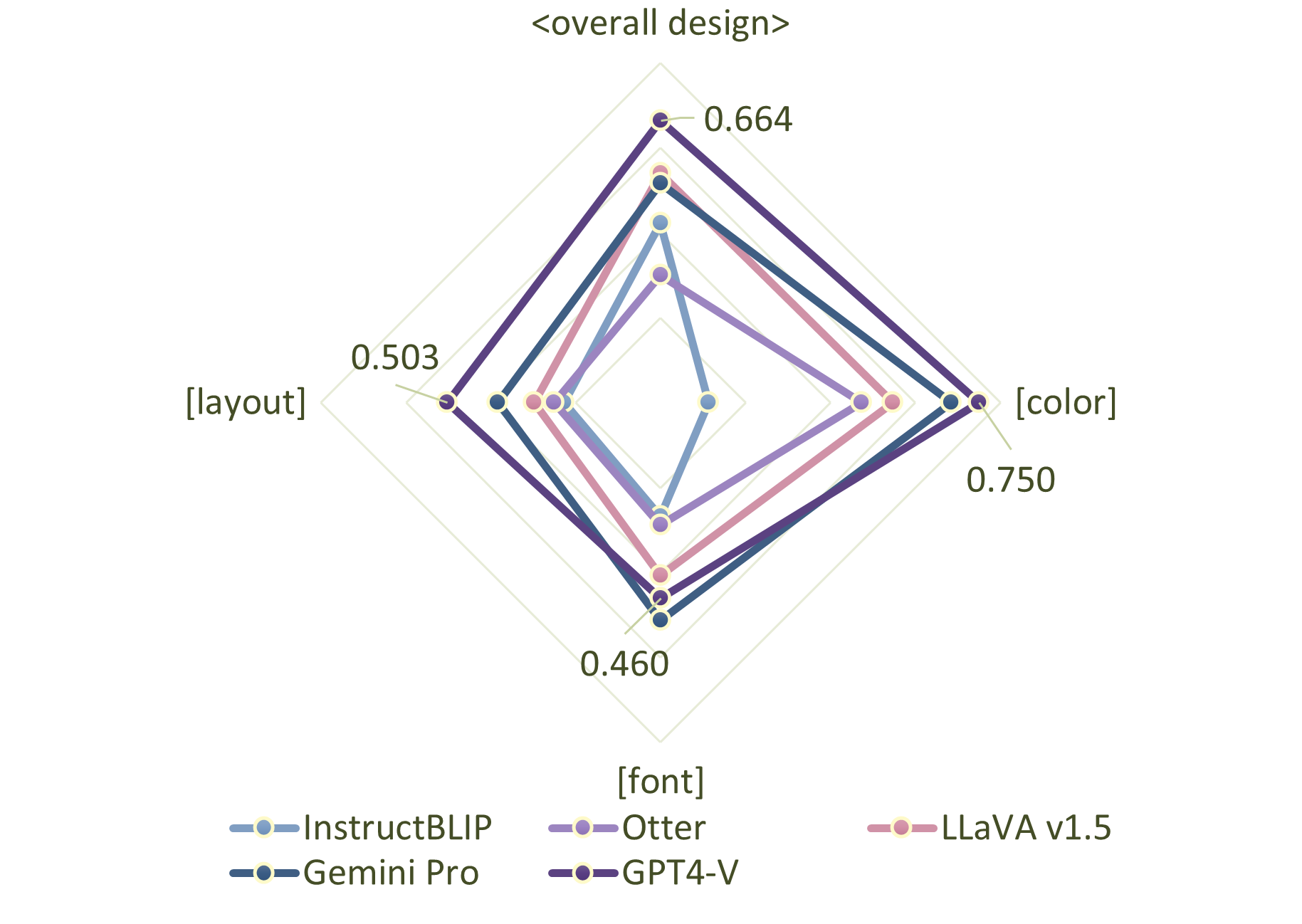}
\caption{The performance of 5 MLLMs at overall design level and design element level (color, font and layout) in DesignProbe.}
\label{fig:radar}
\end{figure}

\begin{figure*}[htb]
\centering
\includegraphics[scale=0.35]{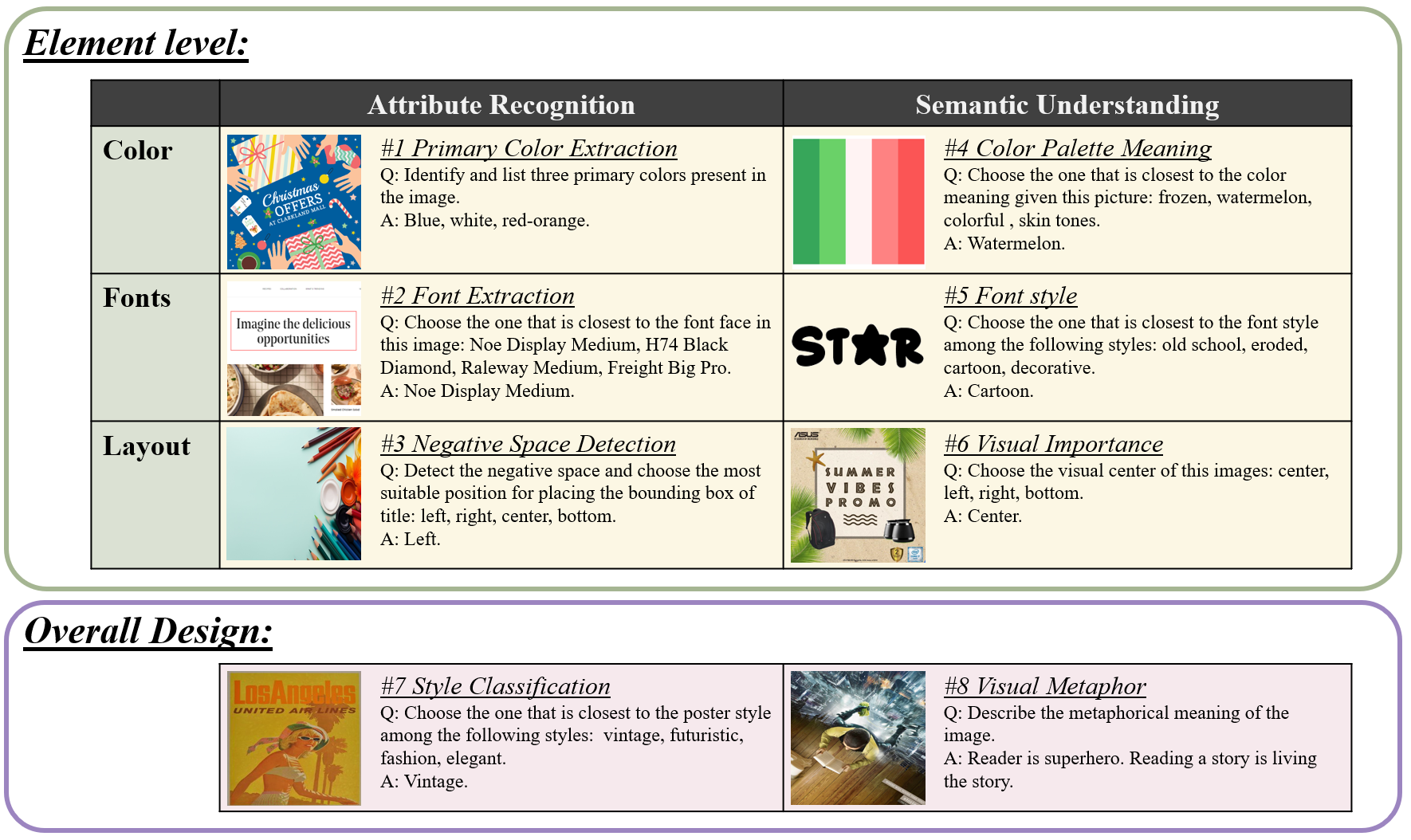}
\caption{Overview of our benchmark. It comprises a total of eight tasks to evaluate the proficiency of MLLM in design. The assessment occurs on two distinct levels: the element level and the overall design level. At the element level, it focus on three fundamental design components: color, font and layout. For each, both visual and semantic aspects are included. Each task is presented with an example.
}
\label{fig:pipeline}
\end{figure*}

\section{Introduction}

Graphic design is essential for our daily experiences, appearing everywhere from movie posters to slides. A well-executed graphic design typically achieves dual-level harmony, weaving together fine-grained design elements such as color, font, and layout with the overall design among different types of elements~\cite{huang2023survey}. 
The fine-grained elements should not only stand alone their own aesthetic principles but also contribute to the overall design harmony. Take the basic design element as examples, colors in design need to have contrast and cooperation between them, which offering clarity and charm, and also require to align with the overall mood and style as well~\cite{color?}.

The complexity inherent in design poses challenges to understanding graphic design. One must first recognize the design elements and then comprehend the overall design. 
The recognition of design elements presents a challenge to existing vision models, for lacking of the design-related data in the pertaining of these models. The different and abstract appearance of these design elements compared with the real word object may add more difficulty in recognition. Comprehending is an equally daunting task as well. They may encounter design tasks for the first time without equipping with the design knowledge, such as the contract and harmony of colors, the different clarity and symbolism carried by different font, and the purposeful arrangement within layout.
Furthermore, there is currently limited insight into the performance of AI systems in understanding design, leaving an area ripe for exploration and advancement.

Recent advancements in the field of Multimodal Large Language Models (MLLMs)~\cite{li2023blip,dai2305instructblip}, especially GPT-4 Vision~\cite{achiam2023gpt}, have demonstrated extraordinary capabilities in a wide range of image-to-text tasks. 
These models perform inspiring  not only in visual recognition tasks such as object detection~\cite{zhu2020deformable} but also in semantic reasoning tasks, including commonsense reasoning~\cite{fu2023mme} and answering college exam questions~\cite{yue2023mmmu} under a zero-shot setting. Inspired by these advancements, we introduce \textit{DesignProbe}, a benchmark designed to explore the performance of MLLMs in the field of design.

In practice, a comprehensive design task commonly need to combine of both recognition and understanding. Building upon the framework delineated in the comprehensive design survey by ~\cite{huang2023survey}, we categorize the design tasks into two distinct levels: the design element level and the overall design level. The former level focus on the detailed dimension of design and different type of elements are considered separately. In this level, our focus is centered on three fundamental components: color, typography and layout. For each, we consider both the dimensions of aesthetic harmony and semantic conveyance. While in overall design level, we focus on the overall feel of the whole design, usually consider different elements together.
To this end, we orchestrate a comprehensive set of eight tasks, the details of which are illustrated in Figure\ref{fig:pipeline}. For evaluation, we apply GPT-4 to evaluate results automatically, which gains similar performance with human annotators but is more stable and cheaper than human. 

Besides the evaluation of existing baseline models, we delve into analytical studies to explore how prompt effect MLLMs performance. 
In order to investigate the variance in model performance with different system prompts and the prompt refinement capabilities of various LLMs, we design a experiment by employing multiple LLMs to rewrite the questions in benchmark. We find that the model with better performance in original task appears to be more robust under different prompts. Besides, refinements using the corresponding LLMs of MLLMs consistently lead to performance improvements.

Moreover, design knowledge is essential as training of MLLMs lacks design task related data.
To equip MLLMs with design knowledge in simple methods, we build up experiments to introduce additional information to prompt in two distinct types: textual and visual information. The experiment results indicate that both types of information are effective, and the direct addition of visual examples leads to substantially greater enhancements in performance compared to textual descriptions alone.

Our main contributions are listed as follow:
\begin{itemize}
    \item To the best of our knowledge, we are the first to conduct a detailed and comprehensive benchmark of design understanding for MLLMs. To facilitate this evaluation, we have curated and re-annotated multiple datasets, and introduce a new dataset for the recognition in layout
    %background layout 
    to enrich the scope of assessment. We test 9 multimodal LLMs, including GPT-4 Vision, Gemini Pro Vision. For evaluation, we employ GPT-4 to measure the distance between ground truth and model outputs. 
    % employing GPT-4 to do automatic evaluation. 
    \item We conduct multiple experiments to refine prompts within the benchmark along with two dimensions: Firstly, to explore the variance of different MLLMs and the prompt refining capability of different LLMs, we conduct a experiments by using LLMs to rewrite the questions.
    Experiment results show the robustness of better performance models under different prompts and the efficiency when refining prompt using MLLM's own LLM.
    Secondly, we incorporate the supplementary knowledge about design tasks to the questions in both text and image types. Experiment results show that adding image type of information directly results in a higher performance gain compared to text. 
    % Both textual and image format of additional knowledge is tested, shows adding image directly can achieve much better performance than text.
    % Experimental results show that the better model usually have better robust in different prompts and it is always effective to refine prompt by its own based LLMs.
    % \item To delve deep into equipping MLLMs with design knowledge, we propose an experiment to add additional information to questions by two format: text descriptions and image examples. Experiment results can reveal that adding image examples directly can achieve much more performance gain than text.
\end{itemize}

\section{Related Work}
\subsection{Graphic Design}
% representation, $understanding$ and generation
% color, font and layout
In recent years, a growing interest has emerged in graphic design. Pioneering researches conducted in this field include tasks such as layout generation~\cite{layouttransformer_2021,layoutdm2023}, color scheme suggestion~\cite{bahng2018coloring}, and font extraction~\cite{ZhaoPG2018}. These tasks can be divided into two distinct levels as outlined by~\cite{huang2023survey}: the element level and the overall design level. At the element level, the goal is to understand or generate a single category of elements separately, while at the overall design level, elements are considered comprehensively. For the element level, there are three basic design elements: color, font, and layout. Color theory has been widely studied, with tasks focusing on both understanding and generation~\cite{bahng2018coloring,qiu2022color}. These tasks may involve building a comprehension task for understanding the color palette or automatically suggesting color palettes for a given design. Fonts are yet another critical element, with research focusing on font extraction~\cite{ZhaoPG2018}, understanding~\cite{dafonts-free}, and generation~\cite{Zhao2018ModelingFI} tasks. For layout elements, there are numerous studies on layout generation, which aim to generate layouts that are visually appealing and convey a clear mood.

Despite the progress in these individual areas, the challenge of capturing and combining features of different design elements remains. \cite{lin2023designbench} is the first to build a design benchmark for text-to-image tasks and show comprehensive results of Dalle-3~\cite{dalle}. The tasks (color, font, layout, and style) in their benchmark are considered and integrated into the task structure of our benchmark. Building upon this foundation, our work presents a more comprehensive image-to-text task structure by adding visual and semantic aspects.

\subsection{Multimodal LLMs}

Multimodal large language models, particularly those that integrate vision and language, such as GPT-4 Vision~\cite{achiam2023gpt} and Gemini~\cite{team2023gemini}, have shown remarkable capabilities. These models excel not only in basic recognition tasks like object identification but also in more nuanced understanding, such as grasping the humor or sentiment behind memes.

The landscape includes not only commercial models like GPT-4 Vision~\cite{achiam2023gpt} but also a burgeoning suite of open-source alternatives. The BLIP series~\cite{li2023blip,dai2305instructblip} and MiniGPT-4~\cite{zhu2023minigpt}, which merge language models with vision encoders, have delivered promising outcomes in tasks that require visual comprehension. LLaVA~\cite{liu2023visual}, on the other hand, has pioneered in generating data that guide models in following multimodal instructions, thereby enhancing their conversational abilities. Furthermore, models like CM3Leon~\cite{yu2023scaling}, DreamLLM~\cite{dong2023dreamllm}, and Emu ~\cite{Emu} have integrated both image understanding and generation into a unified framework. The evolution of multimodal LLMs~\cite{li2023otter,zhang2023llamaadapter,ye2023mplugowl2,Qwen-VL,awadalla2023openflamingo} continues with improvements driven by the incorporation of grounding data, architectural refinements, and other advancements.

This paper examines a collection of these models, spanning both open-source and proprietary frameworks, with an aim to assess their proficiency in interpreting images within the specialized context of graphic design.

% \subsection{Evaluation of Multimodal LLMs}
% benchmarks
The rapid development of multimodal large language models raise the important issue of how to accurately measure their comprehension abilities. Common benchmarks such as captioning ~\cite{chen2015microsoft:cococap,Agrawal_2019_ICCV:nocap} and visual question answering~\cite{goyal2017making,Gurari_2018_CVPR,Hudson_2019_CVPR} have been used to gauge these models' understanding, yet these metrics offer a somewhat limited perspective. To address this limitation, recent efforts~\cite{yu2023mm,fu2023mme,li2023seed,MMBench} have introduced a variety of open-ended evaluation benchmarks that challenge models from multiple dimensions, including cognition and reasoning. Despite these advancements, there remains a gap in evaluating the models' proficiency in specific domains, such as graphic design.

This paper aims to fill this void by proposing a new benchmark tailored to assess the ability of multimodal LLMs in understanding content within the nuanced field of graphic design, thereby providing a more detailed and domain-specific metric for evaluating the capabilities of these sophisticated models.

\section{DesignProbe}
The proposed benchmark comprises a total of eight tasks that evaluate the performance of Multimodal Large Language Models (MLLMs) in design. Below is a detailed introduction to the tasks and the evaluation method.

\subsection{Tasks}

In order to comprehensively assess design capabilities, we prioritize two distinct levels of design: the element level and the overall design level. At the element level, tasks are categorized into two principal aspects: (1) attribute recognition for the visual component and (2) understanding for the semantic dimension. Within each aspect, we focus on three principal design elements: color, font, and layout. Additionally, at the overall design level, we focus on the tasks of style classification and visual metaphor. Figure \ref{fig:pipeline} depicts the framework, which includes a total of eight tasks.

In the element level, we conduct attribute recognition tasks as follows:
\begin{itemize}
    \item \textbf{Task \#1: Color Theme.} The objective is to evaluate the models' ability to identify the primary colors in a design, a critical skill for discerning color harmony and thematic color transitions. We established this task by randomly sampling 50 design instances from Crello~\cite{yamaguchi2021canvasvae}, computing the most frequent colors in a design, and then manually reviewing these examples. For the query structure, we compiled a set of commonly used color palettes, and the models are required to choose their output colors from this predefined selection.
    \item \textbf{Task \#2: Font Extraction.} The task recognizes the font face from a design image where the font is outlined in red. To construct this task, we prompt the model with a single-choice question based on instances randomly sampled from CTXFont~\cite{ZhaoPG2018}.
    \item \textbf{Task \#3: Negative Space Detection.} This task focuses on the detection of negative space, which provides a clear area where elements can be placed without disrupting the visual balance within a design. The model is tasked with analyzing a background image to determine a suitable location for the title. We obtained the background images from Midjourney. After professional designers manually annotated the optimal title locations, the instances were converted into single-choice questions.
\end{itemize}

Semantic understanding tasks are outlined as follows:
\begin{itemize}
    \item \textbf{Task \#4: Color Meaning.} Different combinations of colors can convey different meanings, and certain color palettes may symbolize specific themes or moods. For example, black is rarely found in the color palette associated with ``weddings \& celebrations". We utilize the PAT dataset~\cite{bahng2018coloring} to evaluate this ability. We randomly sampled and manually filtered out examples with ambiguous meanings. The remaining 50 distinct examples were converted into multiple-choice questions.
    \item \textbf{Task \#5: Font Style.}  In addition to recognizing font faces, we expect the model to understand the styles of the given fonts. This is crucial for ensuring that the fonts are consistent with the overall design's mood or theme. We derived the annotations for the fonts and their styles from the Dafonts dataset~\cite{dafonts-free} and transformed them into a single-choice question format.
    \item \textbf{Task \#6: Visual Importance.} Understanding the visual center is essential for layout comprehension and can provide significant feedback for design generation. This task presents the model with a design image and requires it to identify the visual center of the design. We obtained the input images from the Imp-1k dataset~\cite{fosco2020predicting}. Since producing a salience map is challenging for MLLMs and difficult to assess, we categorized the ground truth map into different position descriptions using a 3x3 grid. This task is also formulated as a single-choice question.
\end{itemize} 

For overall design level:
\begin{itemize}
    \item \textbf{Task \#7: Overall Style.} To test the overall design feel of MLLMs, this task asks models to identify the visual style of a given poster from the Poster dataset~\cite{ZhaoSIG2018}. We sampled 50 different examples with a uniform distribution of styles, annotated them with the help of multiple professional designers, and transformed the instances into single-choice questions.
    \item \textbf{Task \#8: Visual Metaphor.} This task delves deeper into understanding the semantic level of design. Visual metaphors often involve using common objects in creative and unfamiliar ways, which can make it challenging to provide the correct caption and the true metaphorical meaning. The designs and explanations were derived from the VisMet dataset~\cite{steen2010method}. After manual filtering, these instances were transformed into open-ended questions.
\end{itemize}
\begin{table}[t!]
    \centering
    \setlength{\tabcolsep}{1mm}
    \resizebox{1\columnwidth}{!}{
    \begin{tabular}{lccccc}
    \toprule
         \multirow{2}{*}{\textbf{Evaluators}} & \multicolumn{4}{c}{Detailed} & \multirow{2}{*}{Acc.} \\
         \cmidrule(lr){2-5}
         & Correct (22) & Partially (7) & Incorrect (20) & Irrelevant (1) \\
         \cmidrule(lr){1-6}
         GPT-3.5-turbo & 95.5 & \textbf{100.0} & 10.0 & 0.00 & 60.0\\
         GPT-4  & \textbf{100.0} & \textbf{100.0} & \textbf{75.0} & \textbf{100.0} & \textbf{90.0}\\
        \bottomrule
    \end{tabular}
    }
    \caption{The results(\%) of GPT-4 and GPT-3.5-turbo as evaluators. Detailed accuracy of each category are shown. The number of cases for each category in the test set is indicated in parentheses following the category. GPT-4 achieves an overall accuracy of \textbf{90\%}, demonstrating its performance to be quite comparable to that of a human evaluator, while significantly reducing labor costs. }
    \label{tab:GPT-4-evaluation}
\end{table}

\subsection{GPT-4 Evaluator}
Although our questions are single-choice, the model still tends to produce open-ended responses. This makes it impractical to compute the performance by simple rules. Therefore, we introduce an automatic evaluation method using GPT-4. Given the question, the golden answer and the MLLMs' generated output, the GPT-4 evaluator is asked to assign a grade by comparing the output with the standard answer. We set the grading scale as [``Correct", ``Partially Correct",``Incorrect", ``Irrelevant"]. Below is the detailed descriptions of each grade:
\begin{itemize}
    \item \textbf{Correct} The style identified in the model's output matches the standard answer perfectly.
    \item \textbf{Partially Correct} This grade applies in two cases:
    
 (1) The style is correctly identified, but the model's answer to the question is wrong. 
 
 (2) The style is incorrectly identified, but the elements of the style are the same as those in the standard answer. 
 \item \textbf{Incorrect} The model's output incorrectly identifies the style of the overall design and the elements within.
\item \textbf{Irrelevant} The model's output does not address the question of style at all.

\end{itemize}

% cite!
% --
To better estimate the performance of the GPT-4 evaluator compared with a human, we build up a test set of 50 questions manually annotated by multiple annotators. We also test the performance of GPT-3.5-turbo with the same prompt. As shown in Table~\ref{tab:GPT-4-evaluation}, GPT-4 achieves an overall accuracy of \textbf{90\%}, which demonstrates that GPT-4 can perform quite similarly to the human evaluator and significantly reduce expensive labor costs.

\begin{table*}[ht]
    \centering
    \setlength{\tabcolsep}{1mm}
    \resizebox{1.75\columnwidth}{!}{
    \begin{tabular}{lccccccccc}
    \toprule
        \multirow{3}{*}{\textbf{Models}} & \multicolumn{6}{c}{\textbf{Element}} & \multicolumn{2}{c}{\textbf{Overall Design}} & \multirow{3}{*}{\textbf{Average}} \\
        \cmidrule(lr){2-7} \cmidrule(lr){8-9}
          & \multicolumn{3}{c}{Recognition} & \multicolumn{3}{c}{Understanding} & \multirow{2}{*}{\#7 Style} & \multirow{2}{*}{\#8 Metaphor} \\
           \cmidrule(lr){2-4} \cmidrule(lr){5-7}
          & \#1 Color & \#2 Font & \#3 Layout & \#4 Color & \#5 Font & \#6 Layout \\
         \cmidrule(lr){1-10}
         random & 20.0 & 25.0 & 25.0 & 25.0 & 25.0 & 25.0 & 25.0 & 0.0 & 21.3 \\
         \cmidrule(lr){1-10}
         % \cmidrule(lr){1-10}
         % QWen-VL-chat & \\
         InstructBLIP & 8.7 & 22.0 & 20.5 & 14.0 & 31.5 & 25.0 & 78.5 & 6.0 & 25.8\\
         MiniGPT-4 & 34.0 & 30.0 & 23.0 & 26.5 & 26.0 & 31.0 & 34.5 & 10.7 & 27.0\\
         Otter & 60.7 & 25.0 & 22.0& 34.0 & 32.5 & 28.0 & 47.5 & 12.7 & 32.8\\
         LLaMA-Adapter v2 & 49.3 & 34.0 & 31.0 & 30.0 & 23.0 & \secondc{} 35.5 & 46.5 & 21.3 &33.8 \\
         
         BLIP-2 & 52.7 & 35.0 &30.0  & 45.5 & 32.0 & \thirdc{} 32.0 & \secondc{} 82.5 & 4.0 & 39.2\\
         mPLUG-Owl2 & 49.3 & \secondc{} 40.5 & \thirdc{} 35.5 & \thirdc{} 54.5 & 33.0 & 28.5 & 66.5 & 18.7 &40.8 \\
         LLaVA v1.5 & \thirdc{} 64.0 & 38.5 & 28.0 & 45.5 & \thirdc{} 43.0 & 31.5 & \thirdc{} 82.0 & \thirdc{} 26.0 &\thirdc{} 44.8 \\
         Gemini Pro Vision & \secondc{} 65.3 & \thirdc{} 39.5 & \secondc{} 50.5 & \secondc{} 71.5 & \firstc{} 63.0 & 26.0 & 70.0 & \secondc{} 33.3 & \secondc{} 52.4 \\
         GPT-4 vision & \firstc{} 72.0 & \firstc{} 43.5 & \firstc{} 55.5 & \firstc{} 78.0 & \secondc{} 48.5 & \firstc{} 45.0 & \firstc{} 87.5 & \firstc{} 45.3 & \firstc{} 59.4\\

        \bottomrule
    \end{tabular}
    }
    \caption{DesignProbe evaluation results (\%) of different MLLMs. All the value in this table is normalized to 1, larger is better. The average values in last column is the average performance of the current MLLM. The table is sorted by average performance. For each column, the highest, the second, and the third highest figures are highlighted by \setlength{\fboxsep}{0pt}\colorbox[RGB]{ 224,174,208 }{purple},  \setlength{\fboxsep}{0pt}\colorbox[RGB]{212, 226, 212}{green} and \setlength{\fboxsep}{0pt}\colorbox[RGB]{ 255,229,229}{pink} backgrounds. }
    \label{tab:mainExp}
\end{table*}

% 对角线 高亮？
\begin{table*}[ht]
    \centering
    \setlength{\tabcolsep}{1mm}
    \resizebox{1.5\columnwidth}{!}{
    \begin{tabular}{lccccccc}
    \toprule
         \textbf{Models} &  based LLM & Ori. & LLaMA2 Re.& Vicuna Re. & GPT-4 Re. & Gemini Re. & std. \\
         \cmidrule(lr){1-8}
         Otter &  MPT & 47.5 & 38.0 & \textbf{52.0} & 33.0 & 48.0 & 7.9\\
         mPLUG-OWL2 & LLaMA2 & 66.5 & \diagonal{} \textbf{77.0} & 76.5 & 70.0 & 74.0 & 4.5\\
         LLaVA & Vicuna v1.5 & 82.0 & 79.5 & \diagonal{} \textbf{83.5} & 80.0 & 81.5 & 1.6\\
         GPT-4 Vision & / & 87.5 & 89.5 & \textbf{90.0} & \diagonal{} 88.0 & \textbf{90.0} &1.2\\
         Gemini Pro Vision& / & 70.0 & 67.5 & \textbf{75.0} & 69.5 & \diagonal{}  72.0 & 2.8\\

        \bottomrule
    \end{tabular}
    }
    \caption{The evaluation results (\%) of different MLLMs using different refined system prompts. \textit{Ori} represents the original questions in DesignProbe. \textit{Re.} is the abbreviation of ``refined". \textit{std.} represent the standard deviation of each row. The corresponding MLLMs with its based LLM are highlighted by \setlength{\fboxsep}{0pt}\colorbox[RGB]{ 217,217,217 }{gray}. The best performance of each MLLMs is in \textbf{bold}. }
    \label{tab:ablation1}
\end{table*}

\begin{figure*}[htb]
\centering
\includegraphics[scale=0.3]{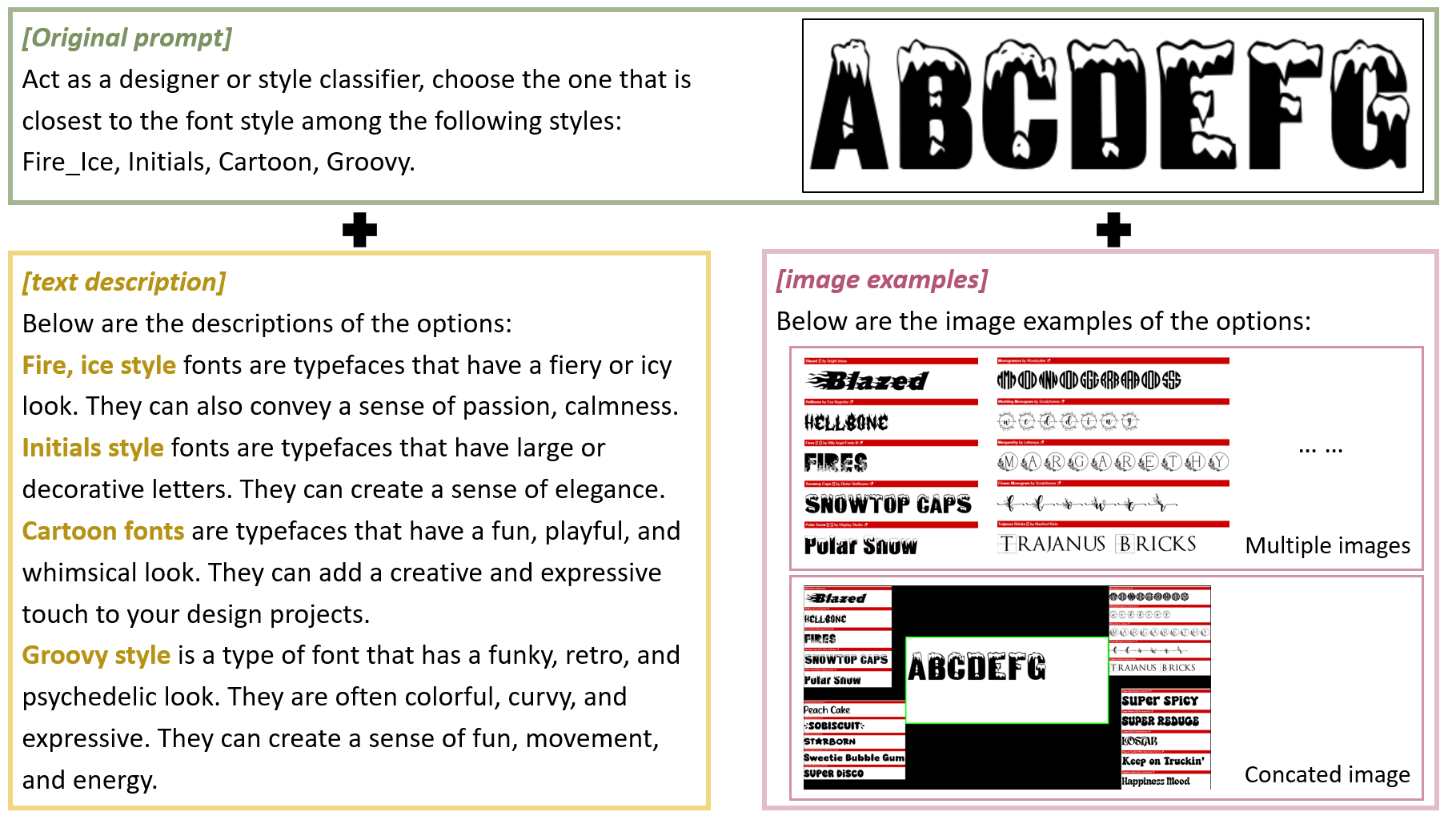}
\caption{The examples of adding example into prompt.
}
\label{fig:ablation2prompt}
\end{figure*}

\begin{figure}[htb]
\centering
\includegraphics[scale=0.5]{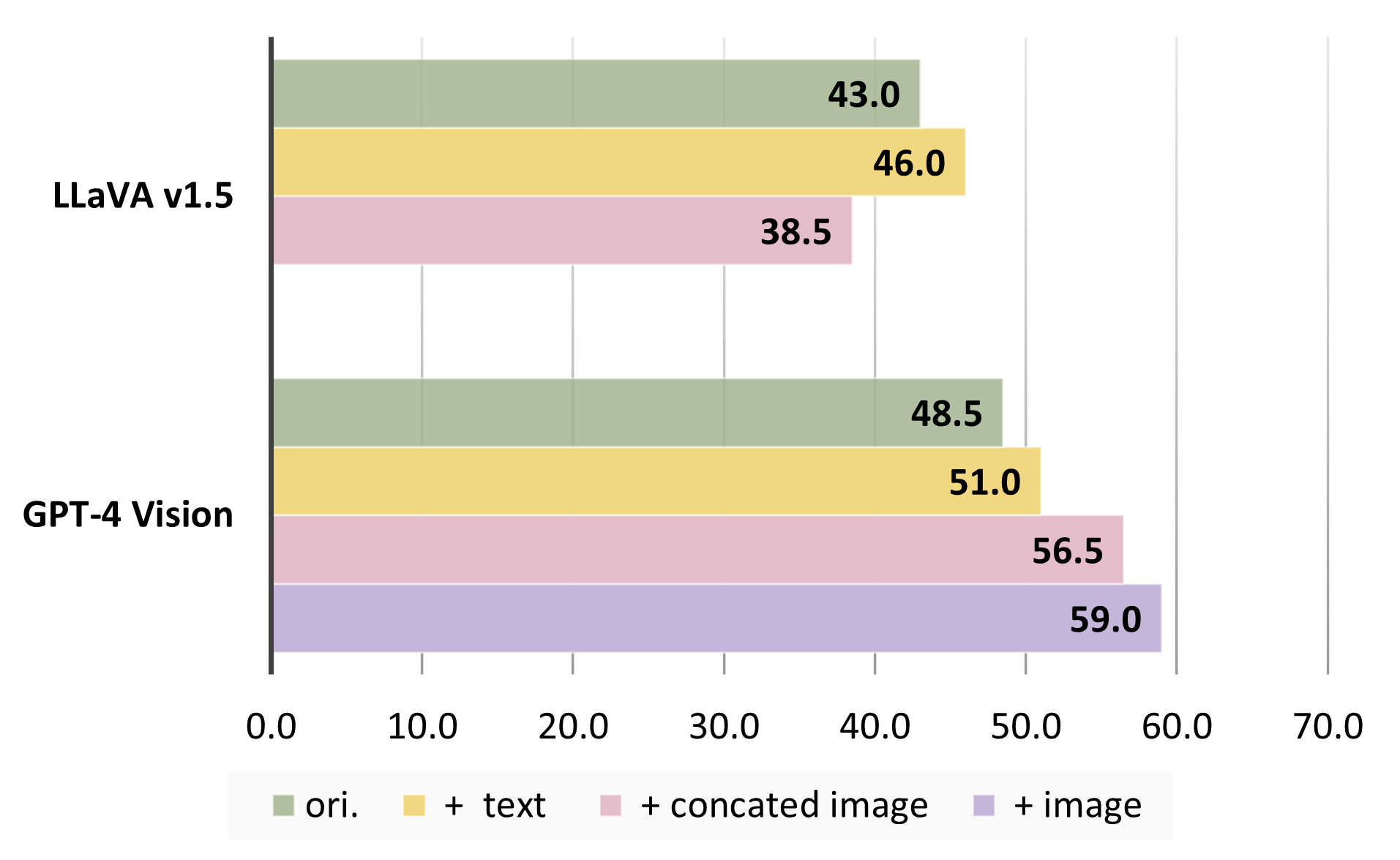}
\caption{The experiment results (\%) of adding additional different types of information to the questions. \textit{Ori} in green represents the performance under original questions in DesignProbe. \textit{ + test} in yellow represents adding text description to the questions. \textit{ + concated image} in pink represents combining multiple image examples into one image due to the unsupportment of multiple images input in LLaVA. \textit{ + image} means adding multiple image examples. 
}
\label{fig:ablation2}
\end{figure}

\begin{figure}[htb]
\centering
\includegraphics[scale=0.4]{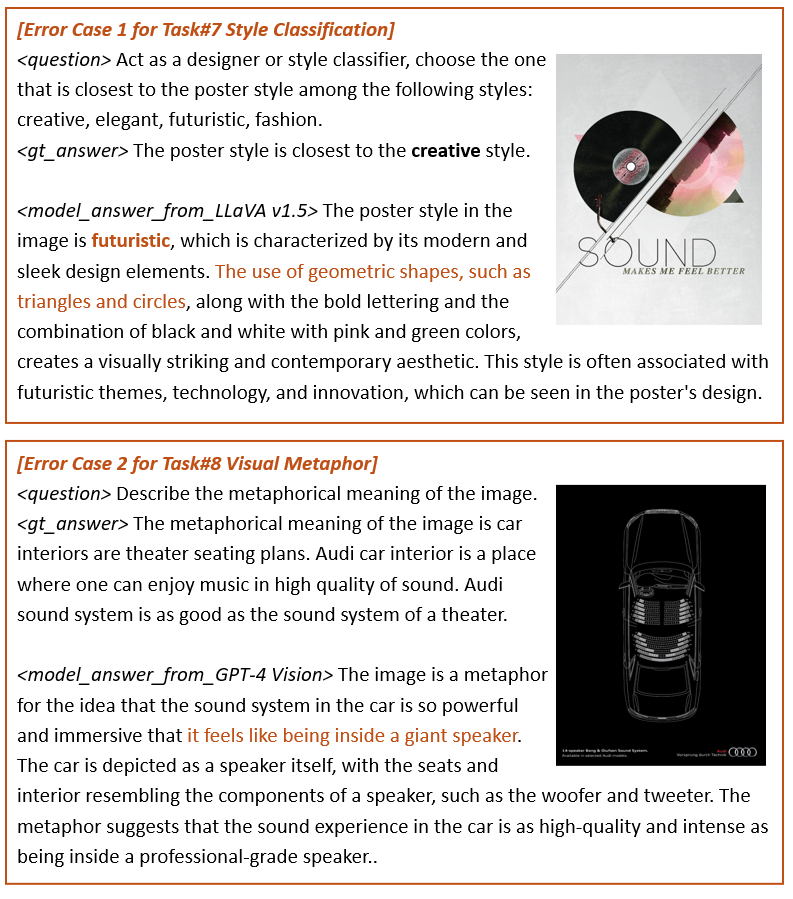}
\caption{Error cases of overall design level tasks. In case 1, the model fails to recognize the creative use of record. In case2, the model fails to recognize the abstract represent of theater seats in car.}
\label{fig:errorCase}
\vspace{-10pt}
\end{figure}

\section{Experiments}

In this section, we conduct extensive experiments on our design benchmark to evaluate a total of nine MLLMs, comprising both open-source and proprietary models.

% exp
To mitigate any positional bias of the correct answer among the various options,  we repeat each question four times with different positions of the correct answer, resulting in a total of 200 questions per task. The results presented in Table~\ref{tab:mainExp} are averaged by position. More detailed results will be provided in the Supplementary Material.

\subsection{Evaluated Models}
% (why we choose these baselines)
% (other img2txt models: VQA, ?) 
% version of each baseline
We evaluate a total of nine MLLMs, selecting the version of each model that demonstrates the best possible performance.

\textbf{LLaVA v1.5}~\cite{liu2023improvedllava} integrates vision and language capabilities through a simple projection layer. We use LLaVA v1.5, which is based on the LLM Vicuna v1.5 13B ~\cite{zheng2023judging}.
% 13B, vicuna v1.5

%Otter, 7B, llama
\textbf{Otter}~\cite{li2023otter} facilitates multimodal in-context instruction tuning, building upon the OpenFlamingo~\cite{awadalla2023openflamingo} model. We evaluate the ``Otter-image-MPT7B" version.

% LLaMA-Adapter-v2, 7B, llama
\textbf{LLaMA-Adapter-v2}~\cite{zhang2023llamaadapter} exclusively employs language data for instruction tuning and establishes a connection between vision and language in a parameter-efficient way. This model is based on LLaMA 7B~\cite{touvron2023llama}.

% MiniGPT-4 v2, LLaMA2, 7B
\textbf{MiniGPT v2}~\cite{chen2023minigptv2} directly projects visual features into LLM feature space using a linear layer and employs unique identifiers for different tasks during training. We use ``MiniGPT v2" version, which is based on LLaMA2 7B~\cite{touvron2023llama2}.

% InstructBLIP, 7B, vicuna v1.1
\textbf{InstructBLIP}~\cite{dai2305instructblip} builds upon BLIP-2\cite{li2023blip} and performs instruction tuning with 26 datasets. We test the model based on Vicuna v1.1 13B.

% mPLUG-OWL2, 7B, llama2
\textbf{mPLUG-OWL2}~\cite{ye2023mplugowl2} utilizes the language decoder as a universal interface to handle different modalities through shared functional modules. We evaluate the ``mplug-owl2-llama2-7b" version.

% BLIP2, FLAN-T5-XXL
\textbf{BLIP2}~\cite{li2023blip} incorporates a Q-Former module to align image features with the LLM token space. This model is based on FLAN-T5-XXL~\cite{chung2022scaling} with a parameter count of 12 billion. 

% GPT-4 Vision
\textbf{Gemini Pro Vision}~\cite{team2023gemini}, \textbf{GPT-4 Vision}~\cite{achiam2023gpt} are evaluated through their respective APIs. Gemini Pro Vision is initially trained using a combination of image and text data. GPT-4 Vision, a large-scale MLLM, performs exceptionally well across various benchmarks~\cite{MMBench,yue2023mmmu}.

% \subsection{Implementations}
% All experiments are conduct on NVIDIA A100 single GPU. 

\subsection{MLLM Performance in DesignProbe}

% overall results: gpt4, gemini better, 
Our benchmark evaluation results are listed in Table~\ref{tab:mainExp}. From these, we summarize our observations into three interesting findings.

\textbf{(1) Overall: Tasks are challenging.}
The highest overall average performance is achieved by GPT-4 Vision at 59.4\%. Despite its significant lead over other baseline models, it is still not enough to meet the passing threshold of 60\%, leaving considerable room for exploration. 

\textbf{(2) Color vs. Font vs. Layout: Models may be more experienced in color than others.} As the results shown in column \textit{\#1 Color} and \textit{\#4 Color} in Table~\ref{tab:mainExp}, we observe an advantage in color-related tasks (with GPT-4 Vision scoring 72.0\% and 43.5\% in color and font). The challenging with font tasks for MLLMs may stem from lacking font-related data during training and instruction tuning. In terms of layout tasks, MLLMs appears to struggle with spatial relationships within design elements, which leads to the performance drop in these related tasks.
% where the color outperform other two type (\#1 Color 72.0\% to 43.5\% in GPT-4 Vision, \#4 Color 54.5\% to 28.5\% in mPLUG-Owl2). The reason why font tasks is difficult for MLLMs may be the lack of font tasks data during training and instruction tuning. As for layout tasks, we find that MLLMs are not sensitive to the position relationships in design, which cause the performance drop in layout tasks.

Interestingly, there is a performance drop between BLIP-2 and InstructBLIP in column \textit{\#1 Color} and \textit{\#4 Color} in Table~\ref{tab:mainExp}. Given that InstructBLIP is essentially BLIP-2 with added instructional tuning, this drop may reveals a trade off between aligning with human preferences and optimizing the model capability.

\textbf{(3) Metaphor: The low performance can be primarily attributed to the models' inability to recognize design objects accurately.} Task\#8 Visual Metaphor is complex as it requires to recognize the abstract design elements and understanding the metaphor they represent. After error analysis, we find there are no instances of 'Correct caption, Wrong reasoning' errors, but an 8\% error rate occurred in cases of 'Correct reasoning, Wrong caption'. This suggests that the main obstacle in metaphor tasks is the models' ability to correctly recognize design objects.

\subsection{Exploration of Prompt Refining}
We conduct multiple experiments to refine prompts within DesignProbe along two distinct dimensions: firstly, we focus on rewriting the task description to enhance clarity and precision without introducing additional information; secondly, we enrich the prompts by providing more contextual and task-related data.

\subsubsection{Prompt Rewriting}
To investigate model response variance to prompts, we design an experiment involving multiple LLMs to refine the original prompt. We then verify the outcomes on \textit{Task\#7 Style Classification}. These LLMs are selected based on the underlying language models of different MLLMs. Evaluating these MLLMs with the refined prompts, we obtain the results shown in Table~\ref{tab:ablation1}. We summarize three key findings as follows:

\begin{itemize}
    \item The better an MLLM performs on a task, the more robust it is to different prompts. For instance, while Otter exhibits significant variance (7.9\%) under different prompts, GPT-4 Vision demonstrates considerable robustness with only a 1.2\% variance. Furthermore, the standard deviation can shed light on MLLM performance, as a small decrease in this variance is often caused by an unsuitable system prompt.
    \item There is always improvement when refining prompts using MLLMs' corresponding base LLMs. For open-source models, employing their own base LLMs yields the best performance. 
    \item Refinement in the language aspect alone generally leads to gains, while other types of refinement may not. To assess the true refinement capabilities of different LLMs, we instruct each LLM to refine questions in their preferred manner using the exact same prompt. Surprisingly, we find that Vicuna consistently performs the best, in contrast to GPT-4. Upon detailed examination of the refined questions, we notice that Vicuna simply rewrites the prompt without adding any text, whereas GPT-4 tends to include additional descriptions of steps for solving the questions. To confirm the impact of these additional texts, we remove them and discover that this action yields the best performance for LLaVA, with an 85\% success rate.
\end{itemize}

\subsubsection{Incorporating Supplementary Information}

The original prompts in DesignProbe are relatively basic and lack detailed task descriptions, potentially leading to confusion for models. To mitigate this, we introduce supplementary information to prompts. Our experiment involves the addition of two types of information: textual descriptions and visual examples. We present the examples of the enhanced prompts for each format in Figure~\ref{fig:ablation2prompt}. To ensure a consistent level of information gain when adding task details, we initially utilize GPT-4 Vision to generate text descriptions for the provided image examples. We then perform minimal manual refinements where necessary. The results of the experiment for \textit{Task\#5 Font Style} are illustrated in Figure~\ref{fig:ablation2}. 

Below are three interesting findings from this experiment:
\begin{itemize}
    \item Incorporating textual descriptions consistently enhances performance. As demonstrated in Figure~\ref{fig:ablation2}, there is an improvement of 3.0\% for LLaVA v1.5 and 2.5\% for GPT-4 Vision.
    \item Adding visual examples directly results in a significantly higher performance gain compared to textual information. For instance, GPT-4 Vision experiences a 10.5\% increase when supplemented with image examples, as opposed to a 2.5\% increase with text alone.
    \item Combining multiple image examples into a single composite image can be a potential workaround for models that only accept a single image input. While LLaVA does not support multiple images, we attempt to merge various image examples into one composite image, drawing inspiration from~\cite{bar2022visual}. However, this approach appears to be counterproductive for LLaVA, as performance decreases from 43.0\% to 38.5\%. Conversely, when applying this method to GPT-4 Vision, the results suggest that such a technique may be beneficial, as it exhibits an opposite effect to LLaVA. This discrepancy leads us to hypothesize that LLaVA may struggle to distinguish between different image examples within a composite image, particularly in tasks requiring spatial recognition, as evidenced by its notably poor performance in layout tasks (\#3, \#6), which is documented in Table\ref{tab:mainExp}.
    
\end{itemize}

\subsection{Error Case Analysis}
In addition to the quantitative results presented earlier, we conduct a thorough analysis of error cases to identify the current limitations of MLLMs in design tasks.

\textbf{Significant Variance Based on the Position of the Correct Option.} We observe that models, such as InstructBLIP and LLaMA-Adapter v2, exhibit considerable variability in performance depending on the position of the correct option among choices A, B, C, and D. We will provide more detailed results in the supplementary materials.

\textbf{Deficiency in Understanding Design Elements.} The models may not be well-acquainted with the tasks or the concepts involved in the tasks, as most tasks within DesignProbe are not covered by the pretraining and fine-tuning datasets of MLLMs. Taking GPT-4 Vision as an example, it exhibits poor performance in \textit{Task\# 5 Font Style} as shown in Table~\ref{tab:mainExp}. However, its performance improves significantly when we supplement it with additional knowledge pertinent to the task. The disparity in performance between tasks involving color and font styles further supports this observation, as evidenced in Table~\ref{tab:mainExp}. Although there are straightforward methods to enhance performance, a more comprehensive investigation is necessary to equip MLLMs with a robust understanding of design principles.

\textbf{Challenges in Recognizing Creative Representations within Design Imagery} Focusing on the general design tasks \#7 and \#8, models frequently struggle to generate precise captions for designs, leading to incorrect responses to subsequent questions. Design objects are often abstract and may diverge significantly from real-world imagery. For instance, as illustrated in Figure~\ref{fig:errorCase}, in case (a), LLaVA fails to recognize the inventive representation of a record and consequently misclassifies the style; in case (b), GPT-4 Vision is unable to identify the theater seats within a car and incorrectly interprets the visual metaphor.

\section{Conclusion}
In this work, we have pioneered the creation of a comprehensive benchmark to assess design capabilities of Multi-Modal Language Models (MLLMs), a first in the field. Our benchmark includes eight tasks across two levels of design complexity: the element level and the overall design level. At the element level, we build up tasks that evaluate both the recognition of visual components and the understanding of semantic content. In each type of tasks, we focus on three fundamental design elements: color, font, and layout. At the overall design level, our tasks include style classification and the interpretation of visual metaphors.
To support these tasks, we have curated and reannotated datasets to align with our novel task framework and introduced a new dataset aimed at negative space detection to extend the benchmark's breadth. Nine MLLMs were put to the test, with GPT-4 serving as the evaluator.

In addition, we experimented with prompt refining across two dimensions. We first introduced a experiment for prompt rewriting using LLMs corresponding to each MLLMs, revealing the robustness of better performance models under different prompts and the efficiency when refining prompt using its LLMs. We also experimented with adding additional design knowledge to prompts in both textual and visual formats, revealing that adding image information can achieve the better performance than text.
This benchmark sets a new standard for evaluating MLLMs and opens the door for future research to expand upon the intersection of design understanding.

%\clearpage
\bibliographystyle{named}
\bibliography{ijcai24}

\end{document}

\maketitle

\section{Prompt of GPT-4 Evaluator}

\begin{figure}[htb]
\centering
\includegraphics[scale=0.3]{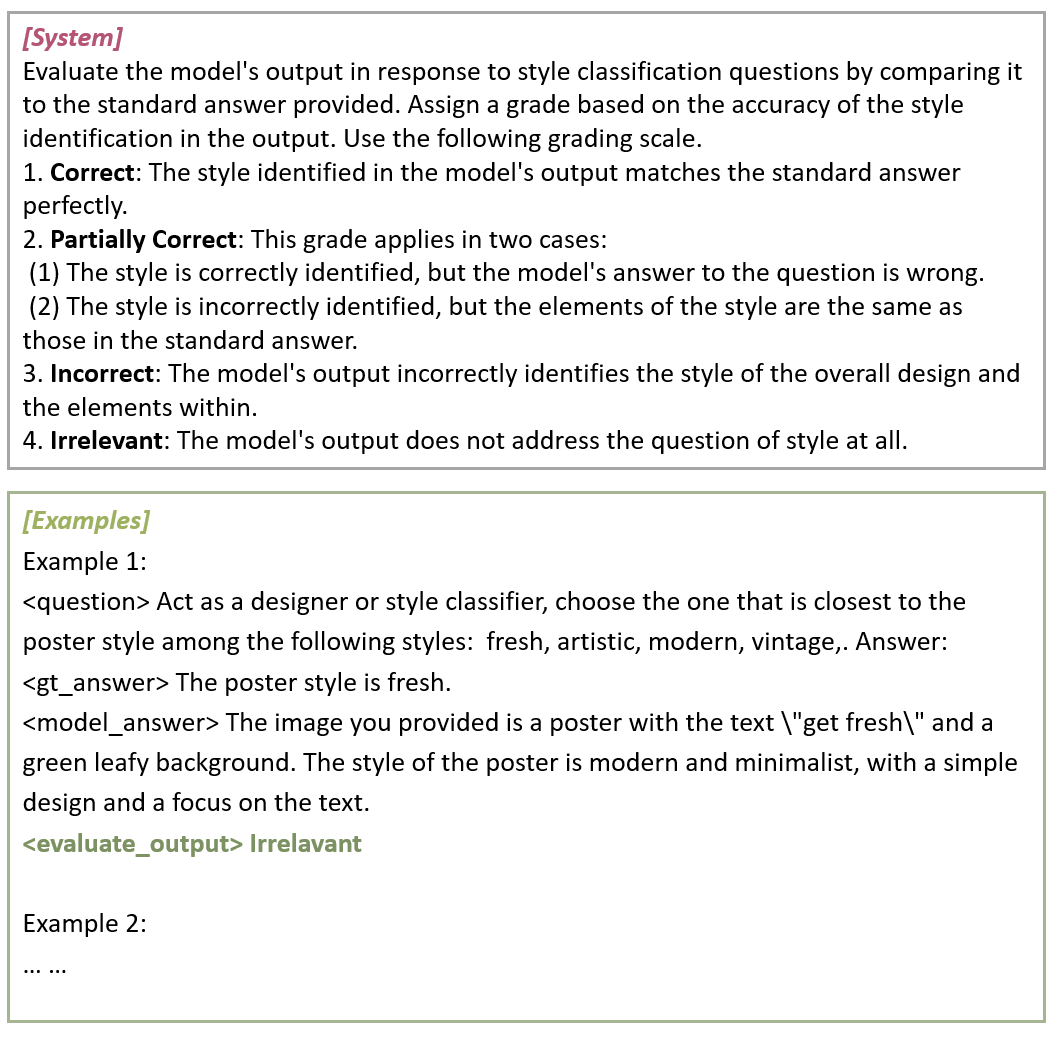}
\caption{The prompt for GPT-4 evaluator. The system prompt provides a detailed description of the evaluation task and grading scales. The in-context learning example prompt comprises questions, golden answers and model outputs.}
\label{fig:gpt4-prompt}
\end{figure}

Figure~\ref{fig:gpt4-prompt} shows the complete prompt of GPT-4 evaluator.

\begin{table}[!h]
    \centering
    \setlength{\tabcolsep}{1mm}
    \resizebox{1\columnwidth}{!}{
    \begin{tabular}{lccccccccccc}
    \toprule
        \multirow{2}{*}{\textbf{Models}} & \multicolumn{5}{c}{\textbf{\#1 Primary Color Extraction}} & \multicolumn{5}{c}{\textbf{\#4 Color Palette Meaning }}\\
        \cmidrule(lr){2-6} \cmidrule(lr){7-11}
         & 0 & 1 & 2 & 3 & avg. & posi 0 & posi 1 & posi 2 & posi 3 & avg. \\
         \cmidrule(lr){1-11}
         random & 48.4 & 43.5 & 7.9 & 0.2 & 0.6 & 25.0 & 25.0 & 25.0 & 25.0 & 25.0 \\
         \cmidrule(lr){1-11}
         InstructBLIP &84.0 & 6.0 & 10.0 & 0.0 & 0.26 & 58.0 & 34.0 & 32.0 & 14.0 & 34.5\\
         MiniGPT4 & 32.0 & 38.0 & 26.0 & 4.0 & 1.02 & 56.0 & 26.0 & 14.0 & 10.0& 26.5  \\         
         Otter & 4.0 & 32.0 & 42.0 & 22.0 & 1.82 & 46.0 & 30.0 & 28.0 & 32.0 & 34.0 \\
         LLama-Adapter v2 & 8.0 & 50.0 & 28.0 & 14.0 & 1.48 & 72.0 & 18.0 & 16.0 & 14.0 & 30.0\\
         BLIP2 & 10.0 & 40.0 & 32.0 & 18.0 & 1.58 & 50.0 & 40.0 & 48.0 & 44.0 & 45.5 \\      
         mPLUG-OWL2 & 8.0 & 46.0 & 36.0 & 10.0 & 1.48 & 58.0 & 56.0 & 60.0 & 44.0 & 54.5\\
         LLaVA v1.5 & 6.0 & 22.0 & 46.0 & 26.0 & 1.92 & 54.0 & 54.0 & 46.0 & 28.0 & 45.5 \\
         Gemini Pro Vision & 8.0 & 22.0 & 36.0 & 34.0 & \underline{1.96} & 66.0 & 70.0 & 80.0 & 70.0 & \underline{71.5} \\
         GPT4-v & 0.0 & 16.0 & 52.0 & 32.0 & \textbf{2.16} & 70.0 & 78.0 & 84.0 & 80.0 & \textbf{78.0} \\
        \bottomrule
    \end{tabular}
    }
    \caption{The detailed results of color tasks in DesignProbe. The highest results in each task are in \textbf{bold} and the second in \underline{underlined}. }
    \label{tab:color}
\end{table}

\section{Breakdown Evaluation Results of Tasks}

We show more detailed evaluation results of different tasks in DesignProbe in Table~\ref{tab:color},~\ref{tab:font},~\ref{tab:layout},~\ref{tab:overallDesign}.

\begin{table}[ht]
    \centering
    \setlength{\tabcolsep}{1mm}
    \resizebox{0.95\columnwidth}{!}{
    \begin{tabular}{lccccccccccc}
    \toprule
        \multirow{2}{*}{\textbf{Models}} & \multicolumn{5}{c}{\textbf{\#2 Font Extraction}} & \multicolumn{5}{c}{\textbf{\#5 Font Style}}\\
        \cmidrule(lr){2-6} \cmidrule(lr){7-11}
         &posi 0 & posi 1 & posi 2 & posi 3 & avg. & posi 0 & posi 1 & posi 2 & posi 3 & avg. \\
         \cmidrule(lr){1-11}
         random & \\
        \cmidrule(lr){1-11}
         InstructBLIP & 52.0& 18.0& 12.0& 6.0& 22.0& 46.0& 30.0& 24.0& 26.0& 31.5\\
         MiniGPT4 & 74.0& 18.0& 18.0& 10.0& 30.0& 50.0& 22.0& 14.0& 18.0& 26.0\\         
         Otter & 58.0& 18.0& 14.0& 10.0& 25.0& 30.0& 34.0& 32.0& 34.0& 32.5\\
         LLama-Adapter v2 &58.0& 22.0& 28.0& 28.0& 34.0& 52.0& 12.0& 12.0& 16.0& 23.0 \\
         BLIP2 & 30.0& 42.0& 36.0& 32.0& 35.0& 32.0& 38.0& 28.0& 30.0& 32.0\\      
         mPLUG-OWL2 & 48.0& 42.0& 40.0& 32.0& \underline{40.5}& 30.0& 44.0& 32.0& 26.0& 33.0\\
         LLaVA v1.5 & 54.0& 52.0& 32.0& 16.0& 38.5& 46.0& 52.0& 42.0& 32.0& 43.0\\
         Gemini Pro Vision & 46.0& 30.0& 38.0& 44.0& 39.5& 78.0& 60.0& 56.0& 58.0& \textbf{63.0}\\
         GPT4-v & 50.0& 36.0& 42.0& 46.0& \textbf{43.5}& 54.0& 42.0& 46.0& 52.0& \underline{48.5}\\

        \bottomrule
    \end{tabular}
    }
    \caption{The detailed results of font tasks in DesignProbe. The highest results in each column are in \textbf{bold} and the second in \underline{underlined}.}
    \label{tab:font}
\end{table}

\begin{table}[ht]
    \centering
    \setlength{\tabcolsep}{1mm}
    \resizebox{0.95\columnwidth}{!}{
    \begin{tabular}{lccccccccccc}
    \toprule
        \multirow{2}{*}{\textbf{Models}} & \multicolumn{5}{c}{\textbf{\#3 Negative Space Extraction}} & \multicolumn{5}{c}{\textbf{\#6 Visual Importance}}\\
        \cmidrule(lr){2-6} \cmidrule(lr){7-11}
         &posi 0 & posi 1 & posi 2 & posi 3 & avg. & posi 0 & posi 1 & posi 2 & posi 3 & avg. \\
         \cmidrule(lr){1-11}
         random & 25.0 & 25.0& 25.0& 25.0& 25.0& 25.0& 25.0& 25.0& 25.0& 25.0\\
         \cmidrule(lr){1-11}
         InstructBLIP & 12.0& 10.0& 36.0& 24.0& 20.5& 34.0& 30.0& 22.0& 14.0& 25.0\\
         MiniGPT4 & 30.0& 20.0& 30.0& 12.0& 23.0& 36.0& 24.0& 26.0& 38.0& 31.0\\         
         Otter & 28.0& 14.0& 22.0& 24.0& 22.0& 30.0& 24.0& 26.0& 32.0& 28.0\\
         LLama-Adapter v2 & 40.0& 28.0& 26.0& 30.0& 31.0& 30.0& 34.0& 46.0& 32.0& \underline{35.5}\\
         BLIP2 & 22.0& 22.0& 32.0& 44.0& 30.0& 20.0& 24.0& 32.0& 52.0& 32.0\\      
         mPLUG-OWL2 & 36.0& 36.0& 42.0& 28.0& 35.5& 28.0& 26.0& 34.0& 26.0& 28.5\\
         LLaVA v1.5 & 62.0& 22.0& 18.0& 10.0& 28.0& 58.0& 20.0& 26.0& 22.0& 31.5\\
         Gemini Pro Vision & 46.0& 54.0& 50.0& 52.0& \underline{50.5}& 30.0& 36.0& 16.0& 22.0& 26.0\\
         GPT4-v & 58.0& 56.0& 50.0& 58.0& \textbf{55.5}& 38.0& 46.0& 54.0& 42.0& \textbf{45.0}\\

        \bottomrule
    \end{tabular}
    }
    \caption{Detailed results of layout tasks in DesignProbe. The highest results in each column are in \textbf{bold} and the second in \underline{underlined}.}
    \label{tab:layout}
\end{table}

\begin{table}[!h]
    \centering
    \setlength{\tabcolsep}{1mm}
    \resizebox{0.95\columnwidth}{!}{
    \begin{tabular}{lccccccccccc}
    \toprule
        \multirow{2}{*}{\textbf{Models}} & \multicolumn{5}{c}{\textbf{\#7 Style Classification}} & \multicolumn{5}{c}{\textbf{\#8 Visual Metaphor}}\\
        \cmidrule(lr){2-6} \cmidrule(lr){7-11}
         &posi 0 & posi 1 & posi 2 & posi 3 & avg. & 0 & 1 & 2 & 3 & avg. \\
         \cmidrule(lr){1-11}
         random & 25.0 & 25.0& 25.0& 25.0& 25.0& 100.0& 0.0& 0.0& 0.0& 0.0\\
         \cmidrule(lr){1-11}
         InstructBLIP & 82.0& 68.0& 78.0& 86.0& 78.5& 94.0& 6.0& 0.0& 0.0& 0.06 \\
         MiniGPT4 & 76.0& 28.0& 20.0& 14.0& 34.5& 70.0& 28.0& 2.0& 0.0& 0.32\\         
         Otter & 46.0& 46.0& 54.0& 44.0& 47.5& 62.0& 38.0& 0.0& 0.0& 0.38\\
         LLama-Adapter v2 &64.0& 24.0& 38.0& 60.0& 46.5& 50.0& 40.0& 6.0& 4.0& 0.64 \\
         BLIP2 & 84.0& 86.0& 84.0& 76.0& \textbf{82.5}& 90.0& 8.0& 2.0& 0.0& 0.12\\      
         mPLUG-OWL2 & 68.0& 56.0& 80.0& 62.0& 66.5& 54.0& 38.0& 6.0& 2.0& 0.56\\
         LLaVA v1.5 & 66.0& 84.0& 92.0& 86.0& \underline{82.0}& 42.0& 42.0& 12.0& 4.0& 0.78\\
         Gemini Pro Vision & 58.0& 64.0& 74.0& 84.0& 70.0& 36.0& 42.0& 8.0& 14.0& \underline{1.00}\\
         GPT4-v & 86.0& 86.0& 88.0& 90.0& 27.5& 16.0& 50.0& 16.0& 18.0& \textbf{1.36}\\

        \bottomrule
    \end{tabular}
    }
    \caption{Detailed results of overall design tasks in DesignProbe. The highest results are in \textbf{bold} and the second in \underline{underlined}.}
    \label{tab:overallDesign}
\end{table}

% \section{Case Study}

% We show two cases for each task from Figure~\ref{fig:ablation2} to Figure~\ref{}.